\documentclass[11pt,a4paper]{article}
\usepackage[hyperref]{acl2017}
\usepackage{times}
\usepackage{latexsym}

\usepackage{url}

\usepackage{graphicx}
\usepackage{booktabs}
\usepackage{tabularx}
\usepackage{amsmath}

\usepackage{multirow}

\aclfinalcopy 

\title{Detecting Off-topic Responses to Visual Prompts}

\author{Marek Rei\\
	    The ALTA Institute\\
	    Computer Laboratory\\
	    University of Cambridge\\
        United Kingdom\\
	    {\tt marek.rei@cl.cam.ac.uk}}

\date{}

\begin{document}
\maketitle
\begin{abstract}
Automated methods for essay scoring have made great progress in recent years, achieving accuracies very close to human annotators. 
However, a known weakness of such automated scorers is not taking into account the semantic relevance of the submitted text.
While there is existing work on detecting answer relevance given a textual prompt, very little previous research has been done to incorporate visual writing prompts.
We propose a neural architecture and several extensions for detecting off-topic responses to visual prompts and evaluate it on a dataset of texts written by language learners.
\end{abstract}

\section{Introduction}

Evaluating the relevance of learner essays with respect to the assigned prompt is an important part of automated writing assessment \cite{Higgins2006,Briscoe2010}.
Existing systems are able to assign high-quality assessments based on grammaticality \cite{Yannakoudakis2011,Ng2013a}, but are known to be vulnerable to memorised off-topic answers which can be a critical weakness in high-stakes testing. In addition, students who have limited relevant vocabulary may try to shift the topic of their answer in a more familiar direction, which most automated assessment systems are not able to capture.
Solutions for detecting topical relevance can help prevent these weaknesses and provide informative feedback to the students.

While there is previous work on assessing the relevance of answers given a textual prompt \cite{Persing2014,Cummins2015,Rei2016b}, very little research has been done to incorporate visual writing prompts. In this setting, students are asked to write a short description about an image in order to assess their language skills, and we would like to automatically evaluate the semantic relevance of their answers.
An intuitive method for comparing multiple modalities is to map them into a shared distributed space -- semantically similar entities will get mapped to similar vector representations, regardless of the information source.
\newcite{Frome2013} used this principle to improve image recognition, by first training separate visual and textual components, and then mapping the images into the same space as word embeddings.
\newcite{Ma2015} performed information retrieval tasks with a related model based on convolutional networks.
\newcite{Klein2015} learned to associate word embeddings to images using Fisher vectors.

In this paper, we start with a similar architecture, based on the approach used by \newcite{Kiros2014a} for image caption generation, and propose modifications that make the model more suitable for discriminating between relevant and irrelevant answers.
The framework uses an LSTM for text composition and a pre-trained image recognition model for extracting visual features. Both representations are mapped to the same space and a prediction is made about the relevance of the text given the image.
We propose a novel gating component that decides which parts of the image should be considered for the current similarity calculation, based on first reading the input sentence. 
Application of dropout to word embeddings and visual features helps increase robustness on an otherwise noisy dataset and assisted in regularising the model. 
Finally, the standard loss function is replaced with a version of cross-entropy, encouraging the model to jointly optimise over batches.
We evaluate on a dataset of short answers by language learners, written in response to visual prompts and our experiments show performance improvements for each of the model modifications.

\begin{figure*}[t]
	\centering
    \hspace{10cm} \includegraphics[width=0.7\linewidth]{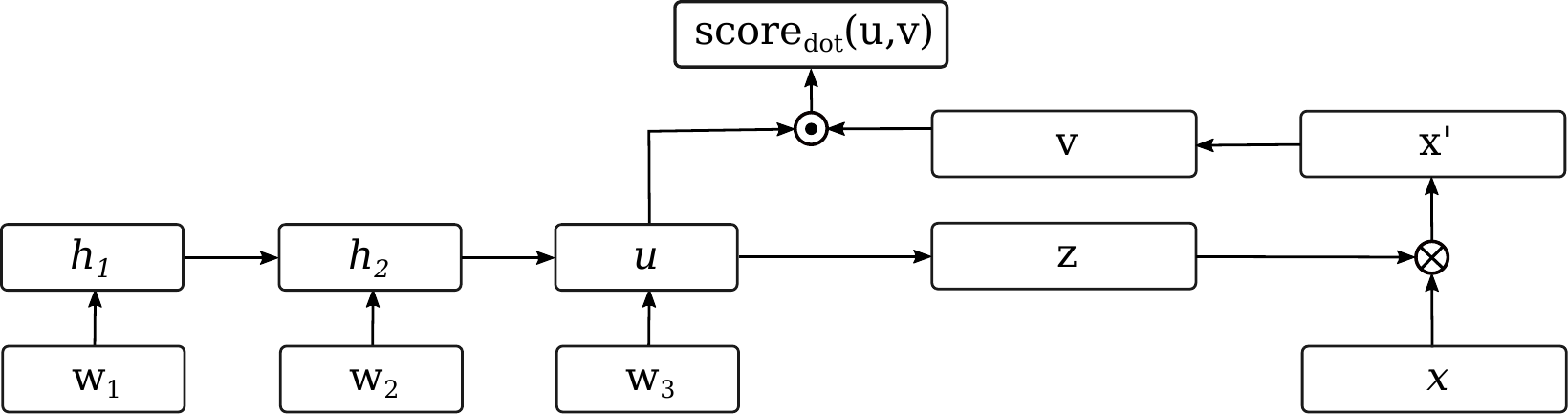}
	\caption{The outline of the relevance detection model. The input sentence and image are mapped to vector representations $u$ and $v$ using modality-specific functions. These vectors are then given to a relevance function which assigns a real-valued score based on their similarity.}
	\label{fig:outline}
\end{figure*}

\section{Relevance Detection Model}

Automated methods for scoring essays and short answers have made great progress in recent years \cite{Yannakoudakis2011,Sakaguchi2015,Alikaniotis2016,Youmna2017}, achieving accuracies very close to human annotators. 
However, a known weakness of such automated scorers is not taking into account the topical relevance of the submitted text.
Students with limited language skills may attempt to shift the topic of the response in a more familiar direction, which automated systems would not be able to detect. 
In a high-stakes examination framework, this weakness could be further exploited by memorising a grammatically correct answer and presenting it in response to any prompt. 
Being able to detect topical relevance can help prevent such weaknesses, provide useful feedback to the students, and is also a step towards evaluating more creative aspects of learner writing.
While there is existing work on detecting answer relevance given a textual prompt \cite{Persing2014,Cummins2015,Rei2016b}, only limited previous research has been done to extend this to visual prompts.
Some recent work has investigated answer relevance to visual prompts as part of automated scoring systems \cite{Somasundaran2015,King2016}, but they reduced the problem to a textual similarity task by relying on hand-written reference descriptions for each image without directly incorporating visual information.

Our proposed relevance detection model takes an image and a sentence as input, and assigns a score indicating how relevant the image is to the text.
Formulating this as a scoring problem instead of binary classification allows us to treat the model output as a confidence score, and the classification threshold can be selected at a later stage based on the specific application.

\newcite{Kiros2014a} describe a supervised method for mapping an image and a sentence into the same space, which allows them to generate similar vector representations for images that have semantically similar descriptions. We base our approach for multimodal relevance scoring on this architecture, and introduce several modifications in order to adapt it to the task of discriminating between relevant and irrelevant textual answers.

The outline of our framework can be seen in Figure \ref{fig:outline}. 
The input sentence is first passed through a Long Short-Term Memory (LSTM, \newcite{Hochreiter1997}) component, mapping it to a vector representation $u$. 
The visual features for the input image are extracted using a model trained for image recognition. The visual representation is then conditioned on the input sentence and mapped to a vector representation $v$. Both $u$ and $v$ are given as input to a function that predicts a confidence score for the answer being relevant to the image.
In the next sections we will describe each of these components in more detail.

\subsection{Text Composition}
\label{sec:text}

The input to the text composition component is a tokenised sentence. We first map these tokens to an embedding space, resulting in a sequence of vector representations:

\begin{equation}
[w_1, w_2, ..., w_N]
\end{equation}

Next, we apply dropout \cite{Srivastava2014a} to each of the word embeddings in the sentence. 
Dropout is a method of regularising neural networks, shown to provide performance imrovements. Neuron activations in a layer are set to zero with probability $p$, preventing the model from excessively relying on the presence of specific features. The process can also be thought of as training a randomly constructed smaller network at each training iteration, resulting in a full combination model. At test time, all the values are retained, but scaled with $(1-p)$ to compensate for the difference. 
While dropout is commonly applied to weights inside the network \cite{Tai2015a,Zhang2015a,Kalchbrenner2015,Kim2016}, there is also some recent work that deploy dropout directly on the word embeddings \cite{Rockt2015,Chen2016}. 
The relevance scoring model needs to handle texts from different domains, including error-prone sentences from language learners, and dropout on the embeddings allows us to introduce robustness into the training process.

We use an LSTM component for processing the word embeddings, building up a sentence representation. 
It is similar to a traditional recurrent neural network, with specialised gating functions that allow it to dynamically decide which information to carry forward or forget.
The LSTM calculates a hidden representation at word $n$ based on the current word embedding and the previous hidden representation at time step $n-1$:

\begin{equation}
h_n = LSTM(w_n, h_{n-1})
\end{equation}

\noindent The last hidden representation $h_N$ is calculated based on all the words in the sequence, thereby allowing the model to iteratively construct a semantic representation of the whole sentence. We use this vector $u = h_N$ to represent a given input sentence in the relevance scoring model.
Since word-level processing is not ideal for handling spelling errors in learner texts, future work could also investigate character-based extensions for text composition, such as those described by \newcite{Rei2016a} and \newcite{Wieting2016}.


\subsection{Image Processing}
\label{sec:image}


In order to map images to feature vectors, a pretrained image recognition model is combined with a supervised transformation component.
We make use of the BVLC GoogLeNet image recognition model, which is based on an architecture described by \newcite{Szegedy2015} and provided by the Caffe toolkit \cite{Jia2014}. The GoogLeNet is a 22-layer deep convolutional network, trained on ImageNet \cite{Deng2009} data to detect 1,000 different image classes. 

An input image is passed through the network and a probability distribution over the possible classes is produced. Instead of using the output layer, we extract the neuron activations at the second-to-last layer in the network -- this takes advantage of all the visual feature processing on various levels of the network, but retains a more general distributed representation of the image compared to using the output layer. 
Similarly to the word embeddings in textual composition, we apply dropout with probability $p$ directly to the image vectors -- this introduces variance to the otherwise limited training data, and prevents the model from overfitting on specific features.

The previous process maps the image to a 1024-dimensional vector $x$, which contains useful visual information but is not optimised for the relevance scoring task.
We introduce a gating component which modulates the image vector, based on the textual vector representation from the input sentence. A vector of gating weights is calculated as a nonlinear weighted transformation of the sentence vector $u$:

\begin{equation}
z = \sigma (u W_z + b_z)
\end{equation}

\noindent where $W_z$ is a weight matrix, $b_z$ is a bias vector, and $\sigma()$ is the logistic activation function with values between $0$ and $1$.
A new image representation $x'$ is then calculated by applying these element-wise weights to the visual vector $x$:

\begin{equation}
x' = z * x
\end{equation}

\noindent where $*$ indicates an element-wise multiplication. This architecture allows the model to first read the input sentence, determine what to look for in the corresponding image, and block out irrelevant information in the image vector. We also disconnect the backpropagation between vector $u$ and the gating weights $z$ -- this forces the model to optimise $u$ only for score prediction, leaving $W_z$ and $b_z$ to specialise on handling the gating.

Finally, we pass the image representation through a fully connected non-linear layer -- this allows the model to transform the pre-trained GoogLeNet space to a representation that is specialised for relevance scoring:

\begin{equation}
v = tanh(x' W_x)
\end{equation}

\noindent where $W_x$ is a weight matrix that is optimised during training, and $v$ is the final image vector that is used as input to the relevance scoring component.

\subsection{Scoring and optimisation}
\label{sec:scoring}


Based on vector representations for the input sentence ($u$) and image ($v$) we now want to assign a score which indicates how related they are. 
\newcite{Kiros2014a} used the cosine measure as the similarity function -- it measures the angle between two vectors, returning a value in the range $[-1,1]$, and is commonly used for similarity calculations in language processing:

\begin{equation}
score_{cos}(u, v) = cos(u, v) = \frac{u v}{|u| |v|}
\end{equation}

The model can then be optimised to predict a high score for image-sentence pairs where the image and sentence and related, and a low score for randomly constructed pairs.
The loss function is a hinge loss with a margin $m$; if the score difference between the positive and negative example is greater than $m$, then no training is required, otherwise the error is backpropagated and weights are updated accordingly:

\begin{equation}
\begin{split}
Loss_{hinge} = \sum_{i \in I} \sum_{j \in J(i)} max(-score_{cos}(u_i, v_i) \\+ score_{cos}(u_j, v_i) + m, 0)
\end{split}
\end{equation}

\noindent where $I$ is the set of related image-text pairs for training, and $J(i)$ is a set of randomly constructed pairs for entry $i$.
When generating the negative examples, we make sure the resulting set $J(i)$ does not contain any examples with the same image as $i$ -- otherwise the model would accidentally optimise related examples towards a low score.

In this work we propose using an alternative scoring function, in order to help discriminate between the answers.
We first replace the cosine similarity with a dot-product:

\begin{equation}
score_{dot}(u, v) = u v
\end{equation}

Next, we create a scoring function by calculating a probability distribution over the current minibatch of examples:

\begin{equation}
score_{exp}(u_i, v_i) = \\
\frac{\exp(score_{dot}(u_i, v_i))}{Z}
\end{equation}

\begin{equation}
\begin{split}
Z = &\exp(score_{dot}(u_i, v_i)) \\&+ \sum_{j \in J(i)} \exp(score_{dot}(u_j, v_i))
\end{split}
\end{equation}

\noindent The model is then optimised for cross-entropy, which is equivalent to optimising the negative log-likelihood:

\begin{equation}
Loss_{ce} = - \sum_{i \in I} log(score_{exp}(u_i, v_i))
\end{equation}

\noindent The transition from cosine to dot-product is required in order to facilitate the new scoring function. In this setting, $score_{exp}(u_i, v_i)$ acts as a softmax layer, requiring the input values to be unbounded for functioning correctly, whereas cosine would restrict values to a range between -1 and 1.

The new scoring function based on softmax encourages the model to further distinguish between relevant and irrelevant images.
While the hinge loss function is also optimised in minibatches, it independently optimises the relevance score of each training pair, whereas softmax connects the scores for all the pairs into a probability distribution. 
When this distribution is optimised using cross-entropy, it specifically focuses more on instances that incorrectly have relatively high scores compared to other pairs in the dataset. In addition, optimising towards a larger score for the known correct example also reduces the scores for all other pairs in the batch.

\section{Evaluation Setup}

\begin{table}[t]
\setlength\tabcolsep{19.5pt}
\begin{tabular}{lrr} \toprule
 & images & sentences \\ \midrule
{\small TRAIN} & 29,000 & 145,000 \\
{\small DEV} & 1,014 & 5,070 \\
{\small TEST} & 1,000 & 5,000 \\ \bottomrule
\end{tabular}
\caption{Number of images and descriptions in the Flickr30k dataset.}
\label{tab:dataset}
\end{table}

Given an image and a text written in response to this image, the goal of the system is to assign a score and return a decision about the relevance of this text.
We evaluate the framework on an experimental dataset collected by 
the English Profile\footnote{http://www.englishprofile.org/},
containing 543 answers written by language learners in response to visual prompts in the form of photographs. 
As part of the instructions, the students were able to select the image that they wanted to write about, and were then free to choose what to write. The length of the collected answers ranges from 1 to 44 sentences.

This dataset contains real-world examples for the task of visual relevance detection, and therefore also proposes a range of challenges.
The answers are provided by students in various stages of learning English, which means the texts contain numerous writing errors. Spelling mistakes prevent the model from making full use of word embeddings, and previously unseen grammatical mistakes will cause trouble for the LSTM composition function.
The students have also interpreted the open writing task in various different ways -- 
while some have answered by describing the content of the image, others have instead talked about personal memories triggered by the image, or even created a short fictional story inspired by the photo. This has led to answers that vary quite a bit in writing style, vocabulary size and sentence length.

Ideally, we would like to train the model on examples where pairs of images and sentences are specifically annotated for their semantic relevance.
However, since the collected dataset is not large enough for training neural networks, we make use of the Flickr30k \cite{Young2014} dataset which contains implicitly relevant pairs of images and their corresponding descriptions.
Flickr30k is an image captioning dataset, containing 31,014 images and 5 hand-written sentences describing each image. We use the same splits as \newcite{Karpathy2015} for training and development; the dataset sizes are shown in Table \ref{tab:dataset}.
During training, the model is presented with 32 sentences and their corresponding images in each batch, making sure all the images within a batch are unique.
The loss function from Section \ref{sec:scoring} is then minimised to maximise the predicted scores for the 32 relevant pairs, and minimise the scores for the $32*32-32=992$ random combinations. 

Theano \cite{Bergstra2010} was used to implement the neural network model. 
The texts were tokenised and lowercased, and sentences were padded with special markers for start and end positions.
The vocabulary includes all words that appeared in the training set at least twice, plus an extra token for any unseen words. 
Words were represented with 300-dimensional embeddings and initialised with the publicly available vectors trained with CBOW \cite{Mikolov2013a}.
All other parameters were initialised with random values from a normal distribution with mean $0$ and standard deviation $0.1$.

We trained for 300 epochs, measuring performance on the development set after every full pass over the data, and used the best model for evaluating on the test set.
The parameters were optimised using gradient descent with the initial learning rate at $0.001$ and the {\small ADAM} algorithm \cite{Kingma2015} for dynamically adapting the learning rate during training.
Dropout was applied to both word embeddings and image vectors with $p = 0.5$.
In order to avoid any outlier results due to randomness in the model, which affects both the random initialisation and the sampling of negative image examples, we trained each configuration with 10 different random seeds and present here the averaged results.

\begin{table*}[t]
\small
\setlength\tabcolsep{10.5pt}
\begin{tabular}{p{0.75cm}p{8.5cm}p{4.5cm}} \toprule
0.65 & In this picture there are lot of people and each one has a different attitude. & \multirow{ 8}{*}{\includegraphics[width=0.3\textwidth, height=60mm]{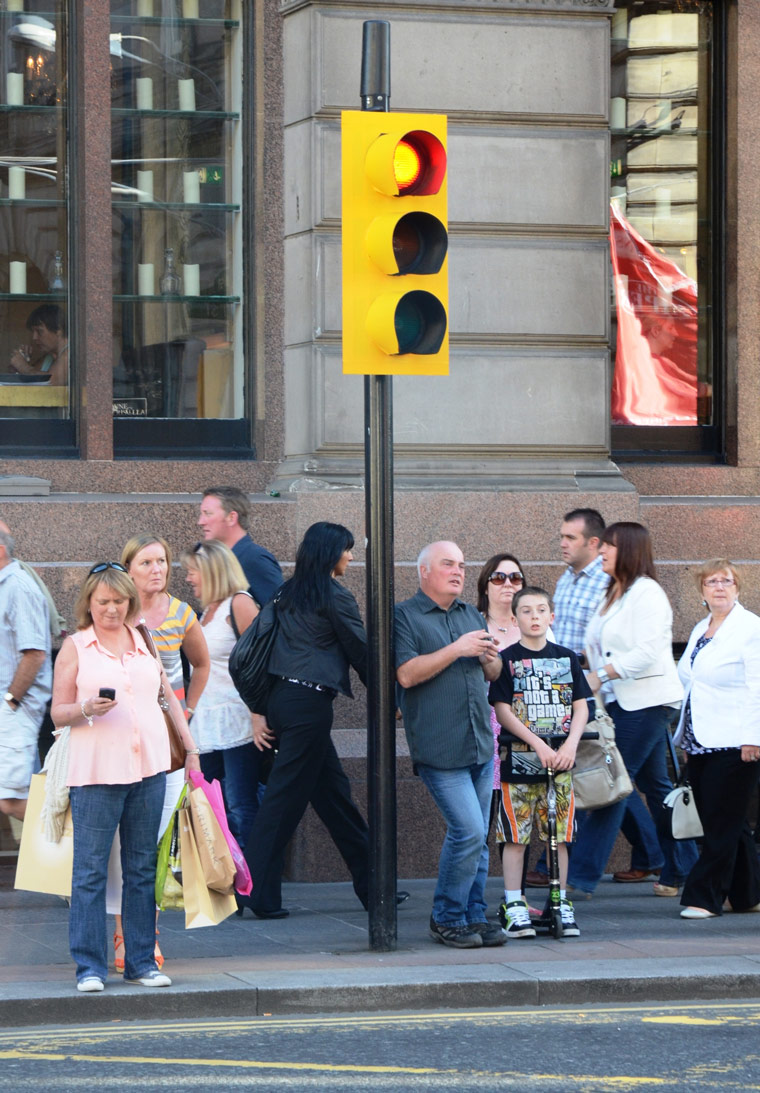}}\\[3ex]
0.81 & In the foreground, people are waiting for the green light in order to cross the street. & \\[3ex]
-2.75 & While a child is talking with an adult about something that is on the other side of the road, instead a women, with lots of bag in her left hand, is chatting with her mobile telephone. & \\[5ex]
0.63 & Generally speaking, the picture is full of bright colours and it conveys the idea of crowded city.& \\[3ex]

-2.38 & Looking at this pictures reminds me of the time I went scuba diving in the sea. & \\[3ex]
-2.16 & It's fascinating, because you are surrounded by water and fishes and everything seems so coulorful and adventurous. & \\[3ex]
-1.40 & Another good part of diving is coming up. & \\[1ex]
-1.70 & You swim to the surface and you see the sunlight coming nearer and nearer until you get out and can breathe "real" air again. & \\[3ex]
\bottomrule
\end{tabular}
\caption{Predicted scores from the best relevance scoring model, given example sentences from the learner dataset and the included photo as a prompt. The first 4 sentences were written in response to this image, whereas the last 4 were written about a different photo.}
\label{tab:realexample}
\end{table*}

\begin{table}[t]
\setlength\tabcolsep{13.5pt}
\begin{tabular}{lccc} \toprule
 & {\small ACC} & {\small AP} & {\small P@50} \\ \midrule
Random & 50.0 & 50.0 & 50.0 \\ \midrule
{\small LSTM-COS} & 68.2 & 71.6 & 81.0 \\
+ gating & 69.6 & 74.6 & 84.4 \\
+ cross-ent & 71.1 & 79.0 & \textbf{92.2} \\
+ dropout & \textbf{75.4} & \textbf{81.9} & 89.8 \\ \bottomrule
\end{tabular}
\caption{Results on the dataset of short answers written by language learners in response to visual prompts. Reporting accuracy, average precision, and precision at rank 50.}
\label{tab:external}
\end{table}

\section{Experiments}
\label{sec:experiments}




We evaluate the visual relevance detection model by training on Flickr30k and testing on the dataset of learner responses to visual prompts.
In order to handle multiple sentences in the written responses, every sentence is first scored individually and the scores are then averaged over all the sentences. 
For every textual answer in the dataset, we create a negative datapoint by pairing it with a random image.
The task is then to accurately detect whether the pair is truly relevant or randomly created, by assigning it high or low relevance scores. 
In order to convert the model output to a binary classification, we employ leave-one-out optimisation -- one example at a time is used for testing, while the others are used to calculate the optimal threshold for accuracy. 
We also report average precision and precision at detecting irrelevant answers in the top 50 returned instances, which measure the quality of the ranking and do not require a fixed threshold.

Results for the different system architectures can be seen in Table \ref{tab:external}. 
The baseline LSTM-COS system is based on the framework by \newcite{Kiros2014a} -- it uses an LSTM for composing a sentence into a vector, calculates the relevance score by finding the cosine similarity between the sentence vector and the image vector, and optimises the model using the hinge loss function. This model already performs relatively well and is able to distinguish between relevant and random image-text pairs with 68.2\% accuracy.

On top of this model we incrementally add 3 modifications and measure their impact on the performance. First, we augment the model with the gating architecture described in Section \ref{sec:image}. The vector representation of the text is used to calculate a dynamic mask, which is then applied to the image vector. This allows the model to first read the sentence, and then decide which parts of the image are more important for the similarity calculation. The inclusion of the gating component improves accuracy by 1.4\% and average precision by 3\%.

Next, we change the scoring and optimisation functions as described in Section \ref{sec:scoring}. Cosine similarity measure is substituted with a dot product between the vectors, removing useful bounds on the score, but allowing more flexibility in the model. 
In addition, the hinge loss function is exchanged for calculating the negative cross-entropy over a softmax. While the hinge loss performs only pairwise comparisons and applies a sharp cut-off, softmax ties all the examples into a probability distribution and provides a more gradual prioritisation for the parameter optimisation. By introducing these changes, the accuracy is again increased by 1.5\% and average precision by 4.4\%.

Finally, we apply dropout with probability $0.5$ to both the 300-dimensional word embeddings in the input sentence and the 1024-dimensional image representation produced by the BVLC GoogLeNet. By randomly setting half of the values to $0$ during training, additional variance is introduced to the available data and the model is becomes more robust for handling noisy learner-generated text. Integrating dropout improves the performance further by 4.3\% and average precision by 2.9\%.


\begin{table}[t]
\setlength\tabcolsep{9.5pt}
\begin{tabular}{lr|rrr} \toprule
 & {\small DEV} & \multicolumn{3}{c}{{\small TEST}} \\
 & {\small ACC} & {\small POS} & {\small NEG} & {\small ACC} \\ \midrule
Random & 16.7 & 0.5 & 0.5 & 16.7 \\ \midrule
{\small LSTM-COS} & 70.8 & 0.7 & 0.0 & 72.6 \\
+ gating & 75.6 & 0.5 & -0.6 & 76.5 \\
+ cross-ent & 82.8 & 5.8 & -5.2 & 83.8 \\
+ dropout & \textbf{87.0} & 5.6 & -3.7 & \textbf{87.4} \\ \bottomrule
\end{tabular}
\caption{Results for different system configurations on the Flickr30k development and test sets. We report accuracy and the average predicted scores for positive and negative examples.}
\label{tab:results}
\end{table}

\begin{figure*}[t]
	\centering
    \includegraphics[width=0.9\linewidth]{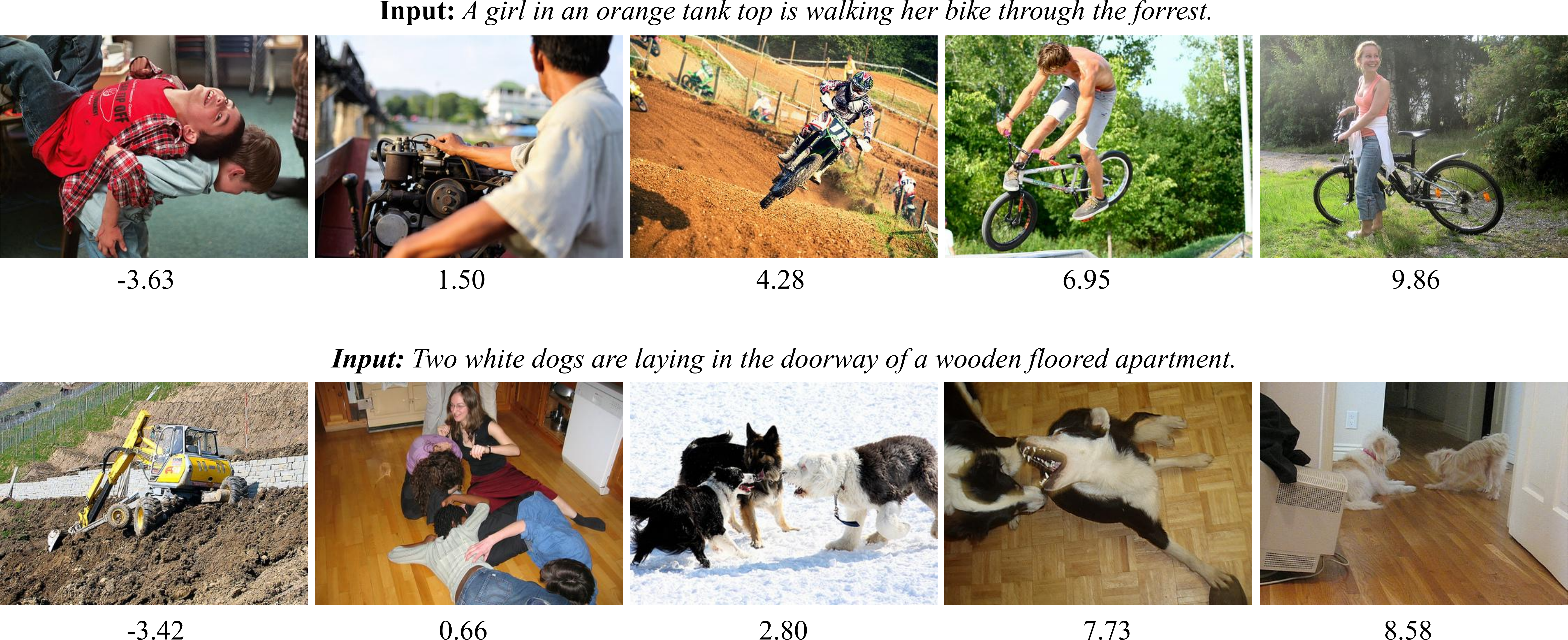}
	\caption{Relevance scores for two example sentences, using the best model from Section \ref{sec:experiments}. Higher values indicate higher confidence in the text being relevant to the image.}
	\label{fig:examples}
\end{figure*}

Table \ref{tab:realexample} contains examples of the predicted scores from the final model, given example sentences written by language learners. 
For most sentences, the model successfully distinguishes between relevant and irrelevant topics, assigning lower scores to the last 4 sentences that describe a different image. However, the model also makes a mistake and incorrectly assigns a low score to the third sentence -- this likely happens due to the sentence being much longer and more convoluted than most examples in the training data, leading the LSTM to lose some important information in the sentence representation.

For comparison, we also evaluate the system architectures on the Flickr30k dataset in Table \ref{tab:results}. 
In this setting, we present the model with a sentence and 6 images from the Flickr30k test set, one of which is known to be relevant while the others are selected randomly. Accuracy is then measured as the proportion of test cases where the model chooses the correct image as the most relevant one.
A random baseline has a 1 in 6 chance of finding the correct image for an input sentence, as there are 5 negative examples for every positive example.
We also report the average scores assigned by the models to positive (relevant) and negative (not relevant) pairs of images and sentences.
As can be seen by the averaged predicted scores in Table \ref{tab:results}, the final system is free to push positive and negative examples apart by a larger margin, increasing the average score difference by an order of magnitude.

\begin{figure*}[t]
	\centering
    \includegraphics[width=0.8\linewidth]{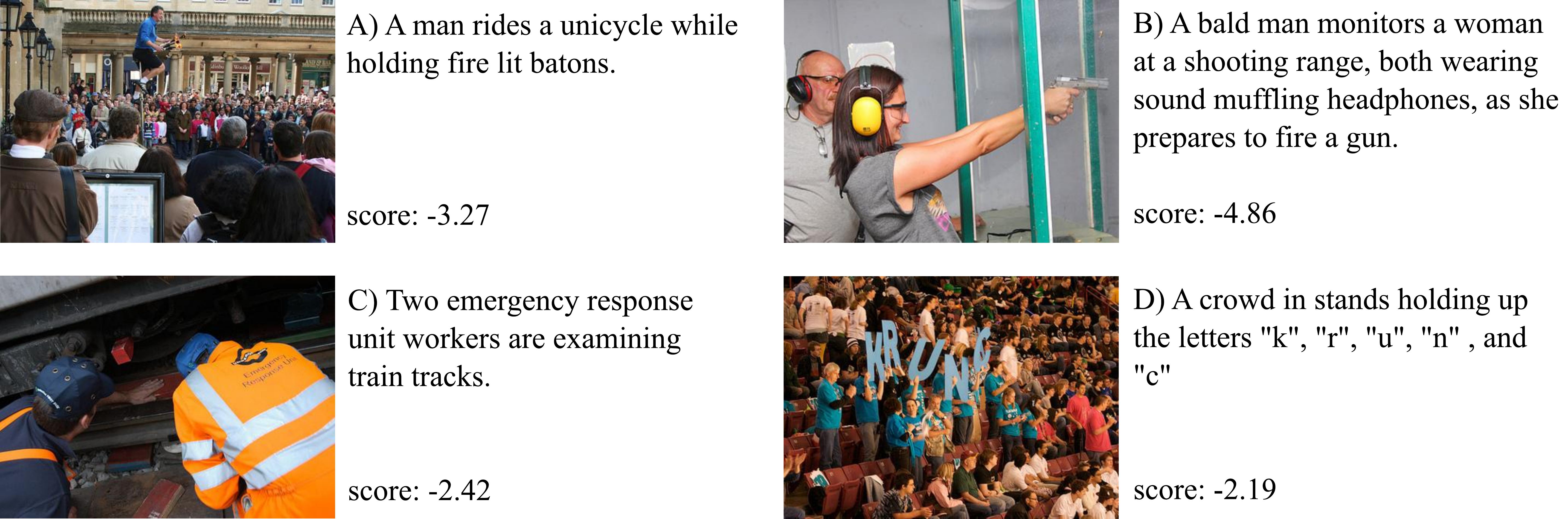}
	\caption{Example valid pairs of images and sentences from the Flickr30k development set where the system incorrectly predicts a low relevance score.}
	\label{fig:errors}
\end{figure*}

\begin{figure*}[t]
	\centering
    \includegraphics[width=0.8\linewidth]{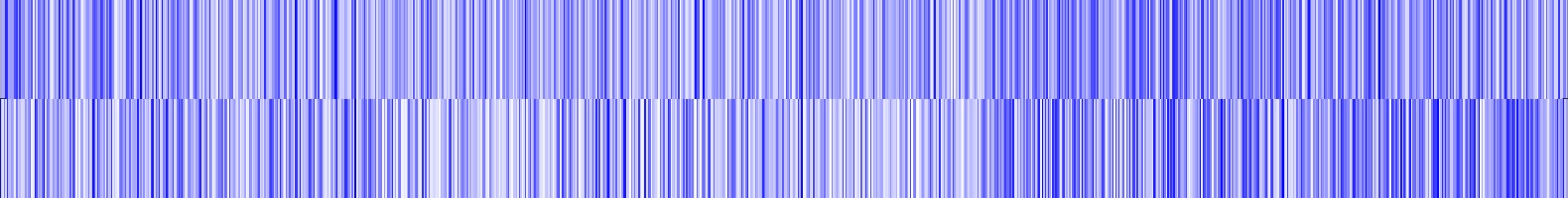}
	\caption{Visualisation of the 1,024 visual gating weights for two example sentences. Lighter areas indicate features where the model chooses to discard the visual information.}
	\label{fig:attention}
\end{figure*}

\section{Analysis}

Figure \ref{fig:examples} contains predicted scores for different images, given example sentences as input.
As can be seen, the system returns high scores when the sentences are paired with very relevant images, and also offers an intuitive grading of relevance. For the first sentence describing an orange shirt and a bicycle, the model has assigned reasonably high scores to other images containing bikes and orange objects. Similarly, for the second sentence the system has found alternative images containing dogs and wooden floors.

In order to analyse the possible weaknesses of the model, we manually examined cases that are difficult for the system.
Figure \ref{fig:errors} contains 4 examples from the Flickr30k development set where a valid image-description pair received a negative score from the relevance model. While a negative score does not necessarily mean an error, as that depends on the chosen threshold, it indicates that the model has low confidence in this being a correct pairing.
The use of rare terms is a source of confusion for the model -- if a word was not used in the training data sufficiently, it will make the relevance calculation more difficult. For example, "unicycle" and "fire lit batons" are relatively rare terms that can cause confusion in example A. In addition, the description mentions only the man, while most of the photo depicts a crowd and a building.

An alternative source of confusion comes from the visual component, with GoogLeNet having more trouble with certain images.
Out of 5,070 image-sentence pairs in the development data, the best model assigned negative scores to 222. Out of those, only 140 had a unique image, indicating that the visual component has more trouble detecting the content of certain unusual images, such as examples C and D, regardless of the textual composition.
Both of these issues represent cases where the model is faced with input that is substantially different from the training examples, and therefore fails to perform as well as possible. This can be remedied by either creating models that are able to generalise better to unseen examples, or by expanding the sources of available training data.

We also analysed the gating component, which is conditioned on the text vector and applied to the image vector. The calculation of the gating weights includes a bias term and a logistic function, which means it could easily adapt to always predicting a vector of 1-s, effectively leaving the image vector unmodified. Instead, we found that the model actively makes use of this additional architecture, choosing to switch off many features in the image vector. Figure \ref{fig:attention} shows a visualisation of the 1024 gating weights for the two example sentences used in Figure \ref{fig:examples}. Values close to $0$ are represented by white, and values close to $1$ are shown in blue. As can be seen, quite a few features receive weights close to zero, therefore effectively being turned off. In addition, the two sentences have fairly different gating signatures, demonstrating that weights are being calculated dynamically based on the input sentence.

\section{Conclusion}

We presented a system for mapping images and sentences into a shared distributed vector space and evaluating their semantic similarity. 
The task is motivated by applications in automated language assessment, where scoring systems focusing on grammaticality are otherwise vulnerable to memorised off-topic answers. 

The model starts by learning embeddings for words in the input sentence, then composing them to a vector representation using an LSTM. In parallel, the image is first passed through a pre-trained image detection model to extract visual features, and then a further supervised layer to transform the representation to a suitable space. We found that applying dropout on both word embeddings and visual features allowed the model to generalise better, providing consistent improvements in accuracy. 

Next, we introduced a novel gating component which first reads the input sentence and then decides which visual features from the image pipeline are important for that specific sentence. We found that the model actively makes use of this component, predicting different gating patterns depending on the input sentence, and substantially improves the overall performance in the evaluations.
Finally, we moved from a pairwise hinge loss to optimising a probability distribution over the possible candidates, and found that this further improved relevance accuracy.

The experiments were performed on two different datasets -- a collection of short answers written by language learners in response to visual prompts, and an image captioning dataset which pairs single sentences to photos. The relevance assessment model was able to distinguish unsuitable image-sentence pairs on both datasets, and the model modifications showed consistent improvements on both tasks. We conclude that automated relevance detection of short textual answers to visual prompts can be performed by mapping images and sentences into the same distributed vector space, and it is a potentially useful addition for preventing off-topic responses in automated assessment systems.

\bibliography{references}
\bibliographystyle{acl_natbib}

\end{document}